\documentclass{article}

\usepackage[preprint, nonatbib]{neurips_2021}

\usepackage[utf8]{inputenc} % allow utf-8 input
\usepackage[T1]{fontenc}    % use 8-bit T1 fonts
\usepackage{hyperref}       % hyperlinks
\usepackage{url}            % simple URL typesetting
\usepackage{booktabs}       % professional-quality tables
\usepackage{amsfonts}       % blackboard math symbols\usepackage{amsfonts}       % blackboard math symbols
\usepackage{nicefrac}       % compact symbols for 1/2, etc.
\usepackage{microtype}      % microtypography
\usepackage{bm}
\usepackage{bbm}

\usepackage{amsmath}
\usepackage{amssymb}
\usepackage{graphicx}
\usepackage{xspace}
\usepackage{physics}
\usepackage{multirow}
\usepackage[table,xcdraw]{xcolor}
\DeclareMathOperator*{\argmax}{arg\,max}

\def\etal{\emph{et al}.}
\newcommand{\veryshortarrow}[1][3pt]{\mathrel{%
 \hbox{\rule[\dimexpr\fontdimen22\textfont2-.2pt\relax]{#1}{.4pt}}%
 \mkern-4mu\hbox{\usefont{U}{lasy}{m}{n}\symbol{41}}}}
 
\DeclareMathOperator{\ent}{\mathcal{H}}
 
\title{Multi-source Few-shot Domain Adaptation}

\author{%
Xiangyu Yue$^{1,}$\thanks{Equal contribution; work done when Zangwei was in Nanjing University.} \And Zangwei Zheng$^{2,}$\footnotemark[1]\And Colorado Reed$^{1}$\And Hari Prasanna Das$^{1}$ \And Kurt Keutzer$^{1}$ \quad Alberto Sangiovanni Vincentelli$^{1}$ 

\\[0.2cm]
$^1$ University of California, Berkeley\\ 

$^2$ National University of Singapore
}

\definecolor{tsneBlue}{rgb}{0.00, 0.42, 0.66}
\definecolor{tsneOrange}{rgb}{1.00, 0.45, 0.15}
\definecolor{tsneGreen}{rgb}{0.00, 0.59, 0.20}
\definecolor{tsneRed}{rgb}{0.85, 0.13, 0.16}
\definecolor{lightPurple}{RGB}{209, 206, 230}
\newcommand{\Gd}{\rowcolor{lightPurple}}
\newcommand{\Gc}{\cellcolor{lightPurple}}

\begin{document}

\maketitle

\begin{abstract}
Multi-source Domain Adaptation (MDA) aims to transfer predictive models from multiple, fully-labeled source domains to an unlabeled target domain. However, in many applications, relevant labeled source datasets may not be available, and collecting source labels can be as expensive as labeling the target data itself. In this paper, we investigate Multi-source \emph{Few-shot} Domain Adaptation (MFDA): a new domain adaptation scenario with limited multi-source labels and unlabeled target data. As we show, existing methods often fail to learn discriminative features for both source and target domains in the MFDA setting. Therefore, we propose a novel framework, termed Multi-Source Few-shot Adaptation Network (MSFAN), which can be trained end-to-end in a non-adversarial manner. MSFAN operates by first using a type of prototypical, multi-domain, self-supervised learning to learn features that are not only domain-invariant but also class-discriminative. Second, MSFAN uses a small, labeled support set to enforce feature consistency and domain invariance across domains. Finally, prototypes from multiple sources are leveraged to learn better classifiers.
Compared with state-of-the-art MDA methods, MSFAN improves the mean classification accuracy over different domain pairs on MFDA by 20.2\%, 9.4\%, and 16.2\% on Office, Office-Home, and DomainNet, respectively. 
\end{abstract}

\section{Introduction}
Deep neural networks have achieved remarkable performance for a variety of computer vision tasks~\cite{he2016deep, long2015fully,redmon2016you, kirillov2019panoptic}. Despite high accuracy, these models have consistently fallen short in generalizing to new domains due to the presence of \textit{domain shift}~\cite{torralba2011unbiased,tzeng2017adversarial,donahue2014decaf}. 
Unsupervised Domain Adaptation (UDA) is a challenging, yet frustratingly practical, setting which aims to transfer predictive models from a single fully-labeled source domain to an unlabeled target domain.
% , which is a common scenario for many applications in autonomous driving, medical imaging, and aerial image analysis~\cite{wang2018deep,zhao2020amulti}. 
UDA methods typically operate by using a task loss on the labeled source samples, as well as additional losses to account for domain shift, such as a discrepancy loss~\cite{long2015learning, sun2016return, sun2017correlation, bousmalis2017unsupervised, zhuo2017deep}, adversarial loss~\cite{tzeng2017adversarial, goodfellow2014generative, liu2016coupled, hoffman2018cycada, russo2018source, sankaranarayanan2018generate}, and reconstruction loss~\cite{ghifary2015domain, ghifary2016deep, bousmalis2016domain}. 

Rather than using only a single labeled source domain, Multi-source Domain Adaptation (MDA)~\cite{mansour2012multiple, duan2012domain,xu2018deep} generalizes this setting by transferring the task knowledge from multiple fully labeled source domains to an unlabeled target domain. 
Each source domain is correlated to the
target in different ways and adaptation involves not only incorporating the combined prior knowledge from multiple sources, but simultaneously preventing the possibility of negative transfer~\cite{ahmed2021unsupervised}. Many MDA methods~\cite{zhao2020multi, xu2018deep, peng2019moment, sebag2019multi, venkat2021your} outperform UDA methods and achieve high accuracy on the target domain by leveraging the abundant explicit supervision in the source domains, together with the unlabeled target samples for domain alignment. 

In many real-world applications, however, getting large-scale annotations even in the source domain is often challenging due to the difficulty and high cost of annotation. 
For example, during the COVID-19 pandemic~\cite{velavan2020covid, cao2020covid}, Harmon \etal~\cite{harmon2020artificial} explored transferring a medical predictive model trained with data from different countries to a target country. However, during the early stages of the pandemic, there were few domain experts that could provide such annotations and even obtaining fully labeled source data was impractical. As another example, each retinal image in the Diabetic Retinopathy dataset~\cite{gulshan2016development} is annotated by a panel of 7+ U.S. board-certified ophthalmologists, with a total group of 54 doctors used for annotation~\cite{gulshan2016development, yue2021prototypical}. Thus it is practically too stringent to assume the availability of abundant labels across domains. 
% Thus, in  many application it is often impractical to obtain a large amount of labeled source data across domains.

In this paper, we explore a new Multi-source Few-shot Domain Adaptation (MFDA) setting that mitigates the need for large-scale labeled source datasets. In MFDA, only a small number of samples in each source domain are annotated while the rest source and target samples remain unlabeled. Many MDA methods seek to learn domain-invariant features by performing some form of distribution alignment~\cite{xu2018deep, peng2019moment, zhao2020multi, zhao2018adversarial, sebag2019multi}, and learn discriminative features by performing supervised task loss on all source domains. In MFDA, however, with a limited number of labels in each source, it is much harder to learn discriminative features for both source and target. Some recent works~\cite{kim2020cross, yue2021prototypical} perform few-shot adaptation with a single source, and in this work, we build upon these contributions and investigate the multiple source scenario.
% there is still no work on MFDA, adapting from multiple sources with limited labels. 

% As the first work on MFDA, in this paper, we 
In the newly proposed MFDA setting, we show that many existing domain adaptation methods do not learn discriminative features for both source and target domains. Therefore, we propose a Multi-Source Few-shot Adaptation Network (MSFAN), which consists of three major components: (i) multi-domain, self-supervised learning (SSL) with feature prototypes, (ii) cross-domain consistency learning that leverages a support-set of labeled and pseudo-labeled samples, and (iii) multi-domain prototypical classifier learning. In the first component, multi-domain prototypical SSL is performed within each domain and each source-target domain pair. The in-domain prototypical SSL aims to learn a well-clustered representation in each domain, while the cross-domain SSL aligns each source domain with the target domain. For the second component, MSFAN builds a support set consisting of the few labeled samples and high-confident unlabeled samples. Based on the support sets, a cross-domain prediction consistency is then enforced to further promote domain-invariant feature learning. Finally, we leverage both the prototypes and the labeled samples from all source domains in order to learn a good classifier for each domain. Mutual information constraints are further enforced on both source and target across all classifiers to learn better domain-invariant and class-discriminative features. 

In summary, our contributions are three-fold. (1) We propose a new domain adaptation task MFDA, adapting to a fully unlabeled target domain from multiple sources with few labels, which is a both practical and challenging generalization of conventional multi-source domain adaptation. (2) To address this challenge, we propose a novel Multi-Source Few-shot Adaptation Network (MSFAN), which learns discriminative and domain-invariant features from multiple domains with only few labels. (3) We conduct extensive MFDA experiments and demonstrate that the proposed method outperforms state-of-the-art MDA methods by large margins across multiple benchmark datasets, with 20.2\%, 9.4\%, and 16.2\% improvement on Office, Office-Home, and DomainNet, respectively. 
%\colorado{let's add concrete results here}

\section{Multi-Source Few-shot Domain Adaptation}

We consider the Multi-source Few-shot Domain Adaptation (MFDA) problem, in which there is one unlabeled target domain and $M$ partially labeled source domains. In the $i$-th source domain, there is small labeled set 
$\mathcal{S}_{i} = \{(\mathbf{x}_i^j, y_i^j)\}_{j=1}^{N_i}$, and a large unlabeled set $\mathcal{S}_{i}^u = \{\mathbf{x}_i^{j,u}\}_{j=1}^{N_i^u}$, both drawn from the source distribution $p_i(\mathbf{x}, y)$ with partial label observation. $N_i$ and $N_i^u$ respectively denote the number of labeled and unlabeled samples in domain $i$, with $N_i \ll N_i^u$. In the target domain, let $\mathcal{T} = \{\mathbf{x}_T^j\}_{j=1}^{N_T}$ denote the target data drawn from the target distribution $p_T(\mathbf{x}, y)$ without label observation, where $N_T$ is the number of target samples. 
We aim to learn an adaptation model that can correctly predict labels of target samples by training on $\{\mathcal{S}_i\}_{i=1}^M$, $\{\mathcal{S}_i^u\}_{i=1}^M$ and $\mathcal{T}$.

\begin{figure*}[t]
 \centering
 \includegraphics[width=5.35in, trim={0.0cm 0cm 0cm 0cm}]{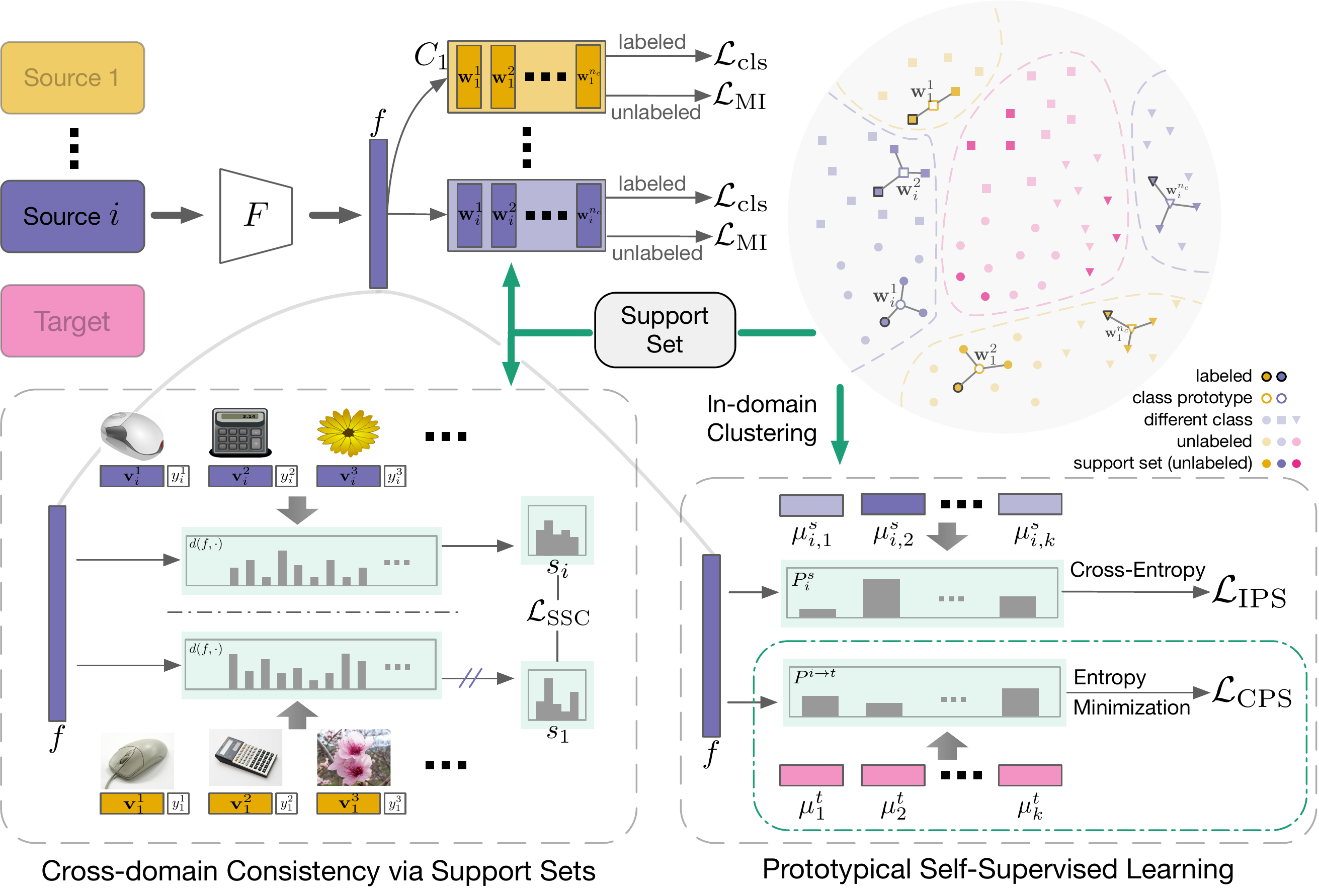}
 %\fbox{\rule[-.5cm]{0cm}{4cm} \rule[-.5cm]{4cm}{0cm}}
 \caption{An overview of the proposed MSFAN framework, which consistes of Multi-domain Prototypical Self-supervised Learning (bottom-right), Cross-domain Consistency via Support Sets (bottom-left), and Prototypical Classifier Learning (top).}
 %\colorado{Let's add more explanation here}}
 \label{fig:framework}
% \vspace{-5pt}
\end{figure*}

Figure~\ref{fig:framework} provides an overview of the Multi-Source Few-show Domain Adaptaion (MSFAN) framework proposed in this paper. It consists of a base model and three major components: Multi-domain Prototypical Self-supervised Learning, Cross-domain Consistency via Support Sets, and Multi-domain Prototypical Classifier Learning. 
Similar to many previous works~\cite{xu2018deep,peng2019moment,wang2020learning,venkat2021your}, the base model of the MSFAN framework consists of a shared feature extractor $F$ and $M$ classifiers $\{C_i\}_{i=1}^M$, one for each source domain. However, instead of a standard linear classifier, each $C_i$ is a cosine similarity-based classifier~\cite{chen2019closer, saito2019semi}. In addition, there is an $\ell_2$ normalization layer between $F$ and $C_i$, which output a feature vector $\mathbf{f}\in \mathbb{R}^d$.

\subsection{Multi-domain Prototypical Self-supervised Learning}
Learning discriminative target features with limited labels per source domain and no labels in the target is a difficult task as the only categorical grounding comes from the few labeled source examples. To address this task, we propose to learn the latent feature-space clustering of each source and target domain, and align clusters with the same category across different domains in a self-supervised manner. 
Specifically, we use a ProtoNCE~\cite{PCL} loss to learn the semantic feature of a single domain as it has been shown to semantically align data across a single source and target domain~\cite{yue2021prototypical}. We further extend it into the multi-source adaptation scenario to learn better discriminative and domain-invariant features across all domains.

\paragraph{In-domain Prototypical Contrastive Learning.} %\colorado{If we have time, and we try to intermix some intuition to what's happening in this subsection. It's hard to follow right now.} 
To learn a well-clustered semantic structure in the feature space, it is problematic to apply ProtoNCE on a mixed dataset with different distributions, because images of different categories from different domains may be incorrectly aggregated into the same cluster. As a result, due to the domain shift between sources and target, we cannot directly apply ProtoNCE to $\bigcup_{i=1}^{M} (\mathcal{S}_i \cup \mathcal{S}_i^u) \cup \mathcal{T}$ as in~\cite{PCL}, and due to the domain shift among different sources, we cannot apply ProtoNCE to the source $\bigcup_{i=1}^{M} (\mathcal{S}_i \cup \mathcal{S}_i^u)$ and target $\mathcal{T}$ separately as~\cite{yue2021prototypical}. Instead, we perform prototypical contrastive learning in each source $\mathcal{S}_i \cup \mathcal{S}_i^u)$ and target $\mathcal{T}$. 

We maintain a memory bank $\bm{V^s_i}$ for each source domain $i$, and a memory bank $\bm{V^t}$ for the target:
\begin{equation}
\bm{V^s_i} = [\mathbf{v}_{i,1}^s, \mathbf{v}_{i,2}^s, \cdots, \mathbf{v}_{i,(N_i+N_i^u)}^s], \,\, \bm{V^t} = [\mathbf{v}_1^t, \mathbf{v}_2^t, \cdots, \mathbf{v}_{N_{T}}^t],
\end{equation}
where $\mathbf{v}_j$ is the feature vector of $\mathbf{x}_j$ which is initialized with $\mathbf{f}_j$ and updated with a momentum $\eta$: %$\mathbf{v}_j \leftarrow m\mathbf{v}_j + (1-m) \mathbf{f}_j$.
\begin{equation}
\mathbf{v}_j \leftarrow \eta\mathbf{v}_j + (1-\eta) \mathbf{f}_j.
\end{equation}
%After each epoch
After a set number of iterations, $k$-means clustering is performed separately on each $\bm{V_i^s}$ to obtain clusters $\bm{C}^s_i = \{C_{i,1}^{s}, C_{i,2}^{s}, \dots, C_{i,k}^{s}\}$, and on the target to obtain target clusters $\bm{C}^t$. Then the normalized prototypes in source $i$ are computed as $\{\mu_{i,c}^s\}_{c=1}^k$, where $\mu_{i,c}^s = \frac{\mathbf{u}^s_{i,c}}{\|\mathbf{u}^s_{i,c}\|}$ with 
$\mathbf{u}_{i,c}^s = \frac{1}{|C^{s}_{i,c}|}\sum_{\mathbf{v}^s_{i,j} \in C^{s}_{i,c}} \mathbf{v}^s_{i,j}$. Similarly, we compute the target prototypes $\{\mu_{c}^t\}_{c=1}^k$. 

To simplify the explanation, we present a set of operations for source $i$ and note that the same operations are applied to all sources and the target.
During the training process, with the feature extractor $F$ and $\ell_2$ normalization layer, we compute a normalized feature vector $\mathbf{f}^s_{i,j} = \ell_2 \circ F(\mathbf{x}_{i,j}^s)$, where $\circ$ represents function composition. Then a similarity distribution vector between $\mathbf{f}^s_{i,j}$ and $\{\mu_{i,c}^s\}_{c=1}^k$ is computed as $P_{i,j}^s = [P^s_{i,j,1}, P^s_{i,j,2}, \dots, P^s_{i,j,k}]$, with 
\begin{equation}
 P^s_{i,j,c} = \frac{\exp ((\mu_{i,c}^s \cdot \mathbf{f}_{i,j}^s - \hat{m}) / \phi) }{\sum_{r=1}^k \exp((\mu_{i,r}^s \cdot \mathbf{f}_{i,j}^s - \hat{m} ) / \phi)},\,\,  \hat{m}= m\cdot\mathbbm{1}_{C_{i,c}^s} (\mathbf{x}_{i,j}^s),
\end{equation}
where $m$ is a margin value inspired by AMS~\cite{wang2018additive}; $\mathbbm{1}_{C_{i,c}^s}(\mathbf{x})$ is an indicator function returning 1 only when $\mathbf{x} \in C_{i,c}^s$; and $\phi$ is a temperature value determining the concentration level of clusters. The in-domain prototypical contrastive loss is then: 
\begin{equation}
 \mathcal{L}_\mathrm{PC} =\sum_{i=1}^{M} \sum_{j=1}^{N_i + N_i^u} \mathcal{L}_{CE} (P_{i,j}^s, c^s_i(j)) + \sum_{j=1}^{N_{T}} \mathcal{L}_{CE} (P_j^t, c^t(j)), 
\end{equation}
where $c_i^s(\cdot)$ and $c^t(\cdot)$ take as input an instance index in a domain and return the cluster index in $\bm{C}^s_i$ and $\bm{C}^t$ respectively. 

Considering the non-deterministic nature of $k$-means algorithm, we perform clustering $R$ times with different number of clusters $\{k_r\}_{r=1}^R$. In MFDA, with prior knowledge of the underling semantic structure (\textit{i.e.} the number of classes $n_c$ is known), we set $k_r = n_c$ for most $r$. Finally, the overall in-domain prototypical self-supervision loss is:
\begin{equation}
 \mathcal{L}_\mathrm{IPS} = \frac{1}{R} \sum_{r=1}^R \mathcal{L}_\mathrm{PC}^{(r)}
\end{equation}

% After? or maybe with
\paragraph{Cross-multi-domain Prototypical Self-supervised Learning.} With the in-domain prototypical learning, the shared network backbone is able to extract discriminative features. To further ensure learning domain-aligned features between the $M$ source domains, and the target domain, we perform cross-multi-domain Prototypical Self-supervised Learning. 

Recently, self-supervised learning methods~\cite{kim2020cross, yue2021prototypical} have been proposed to perform domain alignment between two domains. One trivial extension is combining all source domains first, and then perform domain alignment between $\bigcup_{i=1}^M \mathcal{S}_i \cup \mathcal{S}_i^u$ and $\mathcal{T}$. However, due to the potential domain shift among different sources, aligning the combined source and target would not yield a unified feature distribution. Another trivial extension is aligning each pair of the $M+1$ domains. However, aligning each domain with multiple different domains results in a brittle optimization problem. Moreover, the number of loss terms increases quadratically with the number of source domains. 

In order to address these problems, we propose to perform prototypical domain alignment between each source ($\mathcal{S}_i \cup \mathcal{S}_i^u$) and target ($\mathcal{T}$) pair. For each instance in one source domain, we perform entropy minimization on the distribution similarity vector between its representation and all prototypes in the target domain. 

Specifically, given a feature vector $\mathbf{f}_{i,j}^s$ in source domain $i$, and the prototypes $\{\mu_c^t\}_{c=1}^k$ in the target domain, we first compute the similarity distribution vector $P_{j}^{i\veryshortarrow t} = [P_{j,1}^{i\veryshortarrow t}, \dots, P_{j,k}^{i\veryshortarrow t}]$, with 
% between the feature and each centroid $\mu_j^t$ as:
\begin{equation}
 P_{j,c}^{i\veryshortarrow t} = \frac{\exp(\mu_c^t \cdot \mathbf{f}_{i,j}^s / \tau)}
 {\sum_{r=1}^k\exp( \mu_r^t \cdot \mathbf{f}_{i,j}^s / \tau )},
\end{equation}
in which $\tau$ is a temperature value. To promote confident cross-domain instance-prototype matching,
we minimize the entropy of $P_{j}^{i\veryshortarrow t}$: $H(P_{j}^{i\veryshortarrow t}) = -\sum_{c=1}^k P_{j,c}^{i\veryshortarrow t}\log P_{j,c}^{i\veryshortarrow t}$. 
Note that different from~\cite{yue2021prototypical}, we do not compute and minimize $H(P_{j}^{t\veryshortarrow i})$ on the other direction, since aligning one sample with prototypes in different domains would lead to unstable optimization. The final cross-multi-domain prototypical self-supervised loss is then:
% $H(P_{j}^{t\veryshortarrow i})$ can be computed similarly, and the finally cross-multi-domain prototypical self-supervised loss is: 
\begin{equation}
 \mathcal{L}_\mathrm{CPS} = \sum_{i=1}^M  \sum_{j=1}^{N_s+N_{s}^u} H(P_{j}^{i\veryshortarrow t}),
\end{equation}
% \begin{equation}
%  \mathcal{L}_\mathrm{CPS} = \sum_{i=1}^M \left[ \sum_{j=1}^{N_s+N_{s}^u} H(P_{j}^{i\veryshortarrow t}) + \sum_{j=1}^{N_{T}} H(P_{j}^{t\veryshortarrow i}) \right]
% \end{equation}
and the final objective for the multi-domain prototypical self-supervised learning is then:
\begin{equation}
    \mathcal{L}_\mathrm{MPS} = \mathcal{L}_\mathrm{IPS} + \mathcal{L}_\mathrm{CPS}.
\end{equation}

% \begin{equation}
%  H(P_{j}^{i\veryshortarrow t}) = -\sum_{c=1}^k P_{j,c}^{i\veryshortarrow t}\log P_{j,c}^{i\veryshortarrow t}.
% \end{equation}

\subsection{Cross-domain Consistency via Support Sets}
\label{subsec:support-set}
To further promote domain-invariant and class-discriminative features, we propose to enforce a cross-domain similarity consistency using a support set of labeled and pseudo-labeled data. 
\paragraph{Support Set. } We build a support set $S^{(i)}$ for each source domain $i$. The support samples in the set contains the labeled samples in $S_i$, and unlabeled samples in $S_i^u$ with consistent high-confident predictions across all classifiers. Formally, $S^{(i)}$ is compuated as:
\begin{equation}
 S^{(i)} = \mathcal{S}_i \, \cup \{(\mathbf{x}, y) |\, \mathbf{x} \in \mathcal{S}_i^u; \,\,\forall i', \max \mathbf{p}_{i'}(\mathbf{x}) > t, \argmax\mathbf{p}_{i'}(\mathbf{x})=y \},
\end{equation}
where $\mathbf{p}_i(\mathbf{x})$ is the softmax vector from $C_i$ on $\mathbf{x}$, $y=\argmax\mathbf{p}_i(\mathbf{x})$, and $t$ is a confidence threshold. 

Let $\bm{V}_{\mathcal{S}^{(i)}}$ denote the support representations computed from $S^{(i)}$, then for a scalar-valued distance function $d(\cdot,\cdot)$, the similarity vector between an input vector $\mathbf{f}_j$ and $S^{(i)}$ can be computed as:
\begin{equation}
    \mathbf{s}_{i,j} = \sum_{(\mathbf{v}^s_{i,k}, y_k) \in \bm{V}_{\mathcal{S}^{(i)}}} \left( \frac{d(\mathbf{f}_j, \mathbf{v}^s_{i,k})}{\sum_{(\mathbf{v}^s_{i,r}, y_r)\in \bm{V}_{\mathcal{S}^{(i)}} } d(\mathbf{f}_j, \mathbf{v}^s_{i,r}) } \right) y_k,
\end{equation}
where $y_k$ is the one-hot ground truth label vector associated with the $k$-th representation in $\mathbf{v}_{\mathcal{S}^{(i)}}$. In this paper, we choose $d(a , b)$ to be $\exp(\frac{a\cdot b}{\Vert a\Vert \Vert b\Vert \psi})$, where $\psi$ is a temperature value. 

Given an image $\mathbf{x}_j \in \bigcup_{i=1}^{M} (\mathcal{S}_i \cup \mathcal{S}_i^u) \cup \mathcal{T}$, we want to enforce consistency on its similarity vectors across different source domains by minimizing the cross entropy. Finally, with the similarity vector to another domain $i'$, $\mathbf{s}_{i',j}$, as the soft pseudo-label, the loss can be computed as:% \zangwei{Some problem with sharpening}
\begin{equation}
    \mathcal{L}_\mathrm{SSC} = \sum_{\mathbf{x}_j \in \mathcal{D}}\sum_{1 \leq i  \neq i' \leq M} \mathcal{L}_{CE} (\mathbf{s}_{i,j}, \mathbf{s}_{i',j})
\end{equation}
\subsection{Multi-domain Prototypical Classifier Learning}
MSFAN incorporates a multi-domain prototypical classifier to learn better domain-aligned, class-discriminative features. It accomplishes this through a simple cosine classifier for each source domain $\{C_i\}_{i=1}^M$.
Each cosine classifier $C_i$ consists of weight vectors $\mathbf{W}_{i} = [\mathbf{w}_{i}^1, \mathbf{w}_{i}^2, \dots, \mathbf{w}_{i}^{n_c}]$, where $n_c$ is the number of classes, and a temperature $T$. The output of $C_i$, ${\frac{1}{T} {\mathbf{W}_{i}}^{\intercal}\mathbf{f}}$, is fed into softmax layer $\sigma$ to obtain the final probabilistic output $\mathbf{p}_i(\mathbf{x})=\sigma ({\frac{1}{T} {\mathbf{W}_{i}}^{\intercal}\mathbf{f}})$. Most previous MDA works train the classifier of domain $i$ with only the labeled data in domain $i$. However, with only few labeled samples per source in MFDA, $C_i$ is prone to overfit to $\mathcal{S}_i$. Thus we train each $C_i$ with labeled data from all source domains using a standard cross-entropy loss:
\begin{equation}
    \mathcal{L}_{\mathrm{cls}}^{(i)} = \mathbb{E}_{(\mathbf{x},y)\in \bigcup\mathcal{S}_i} \mathcal{L}_{CE}(\mathbf{p}_i(\mathbf{x}), y)
\end{equation}
One drawback of training each $C_i$ using the same set of all labeled data is that the predictive behavior of each $C_i$ will likely be quite similar, which greatly impairs the ensembling effect of multiple classifiers during test time. To build a classifier set $\{C_i\}$ with more diverse predictive behavior, we desire to make each $C_i$ have slightly higher accuracy on domain $i$ than on other domains. 

Looking closer at a cosine classifier $C$ with weight matrix $\mathbf{W}$, in order for it to have high performance, the $k$-th weight vector $\mathbf{w}^k$ needs to be a representative vector of the corresponding class $k$. To promote a more diverse set of $\{C_i\}$, we directly update $\mathbf{W}_i$ with prototypes computed from the corresponding support set $S^{(i)}$, computed in Sec.~\ref{subsec:support-set}. Specifically, we estimate prototype of class $k$ in domain $i$ as:
\begin{equation}
\label{equ:weight-update}
    \mathbf{\hat{w}}_{i}^k = \frac{1}{|S^{(i)}_k|} \sum_{\mathbf{x}\in S^{(i)}_k} \bm{V^s_i}(\mathbf{x}),
\end{equation}
where $S^{(i)}_k = \{\mathbf{x}| y=k, \forall (\mathbf{x}, y)\in S^{(i)}\}$, $\bm{V^s_i}(\mathbf{x})$ returns the representation of $\mathbf{x}$ stored in the memory bank, and $\mathbf{w}_{i}^k$ is then updated with $\frac{\hat{\mathbf{w}}_{i}^k}{\Vert \hat{\mathbf{w}}_{i}^k \Vert}$ frequently.% \zangwei{weight annotation} 

\paragraph{Classifier-wise Mutual Information} To learn better domain-invariant and discriminative features, we maximize the mutual information between the input and output of each classifier with unlabeled images across all domains. For classifier $C_i$, the mutual information can be written as
\begin{equation}
 % \mathbb{E}[p(y|x;\theta)]
 \mathcal{I}_i(y;\mathbf{x}) = \mathcal{H}(\mathbf{\Bar{p}}_i) - \mathbb{E}_{\mathbf{x}}[\mathcal{H}(p(y|\mathbf{x};\theta_i))], 
\end{equation}
where $p(y|\mathbf{x};\theta_i)$ denotes the output of $C_i$ on $\mathbf{x}$ and $\mathbf{\Bar{p}}_i$ is a prior distribution $\mathbb{E}_\mathbf{x}[p(y|\mathbf{x};\theta_i)]$. We can get the mutual information maximization objective as:
\begin{equation}
    \mathcal{L}_\mathrm{MI} = - \sum_{i=1}^M \mathcal{I}_i(y;\mathbf{x}), \,\,\, \mathbf{x} \in \bigcup_i \mathcal{S}_i^u \cup \mathcal{T}
\end{equation}

\subsection{MSFAN Learning}
The MSFAN learning framework performs multi-domain prototypical self-supervised learning, support-set-based cross-domain similarity consistency, and multi-domain prototypical classifier learning. Together with the classifier update with Eq. \ref{equ:weight-update}, the overall training objective is:
\begin{equation}
    \mathcal{L}_\mathrm{MSFAN} = \mathcal{L}_\mathrm{cls} + \lambda_\mathrm{mps}\cdot \mathcal{L}_\mathrm{MPS} +\lambda_\mathrm{ssc} \cdot \mathcal{L}_\mathrm{SSC} + \lambda_\mathrm{mi} \cdot \mathcal{L}_\mathrm{MI}
\end{equation}
\paragraph{Global Max-Similarity-based Inference} For target data during test time, we propose a new inference method based on the max-similarity across all classifiers. With normalized weights from all classifiers $\mathcal{W}=\{\mathbf{w}_{i}^c | 1\leq i\leq M, 1\leq c\leq n_c\}$, given a test example feature $\mathbf{f}_t$, the most similar weight vector is identified as $\mathbf{w}_{i^*,c^*}=\argmax_{\mathbf{w} \in \mathcal{W}} \mathbf{f}_i^{\intercal} \cdot \mathbf{w}$, and $c^*$ is the final prediction. 

% which can be regarded as the prototype of class $k$. 

\section{Experiments}
\subsection{Experimental Setting}
\paragraph{Datasets.}\label{para:dataset} We evaluate our method (MSFAN) in multi-source few-shot setting on three standard domain adaptation benchmarks, Office~\cite{saenko2010adapting}, Office-Home~\cite{venkateswara2017deep}, and DomainNet~\cite{peng2019moment}. The labeled data in each domain are chosen following~\cite{kim2020cross,yue2021prototypical}, and each domain is in turn regarded as the target domain, while the others in the same dataset are considered as source domains. 
 \textbf{Office}~\cite{saenko2010adapting} is a real-world dataset for domain adaptation tasks. It contains 3 domains (Amazon, DSLR, Webcam) with 31 classes.
Experiments are conducted with 1-shot and 3-shots source labels per class in this dataset. 
\textbf{Office-Home}~\cite{venkateswara2017deep} is a more difficult dataset than Office, which consists of 4 domains (Art, Clipart, Product, Real) in 65 classes. 
Following~\cite{kim2020cross,yue2021prototypical}, we look into the settings with 3\% and 6\% labeled source images per class, which means each class has 2 to 4 labeled images on average.
\textbf{DomainNet}~\cite{peng2019moment} is a large-scale domain adaptation benchmark. Since some domains and classes are noisy, we follow~\cite{saito2019semi,yue2021prototypical} and use a subset containing four domains (Clipart, Painting, Real, Sketch) with 126 classes.
We show results on settings with 1-shot and 3-shots source labels on this dataset.

\paragraph{Implementation Details. }
We use ResNet-101~\cite{he2016deep} (for DomainNet) and ResNet-50 (for other datasets) pre-trained on ImageNet~\cite{russakovsky2015imagenet} as backbones for all baselines and MSFAN. To enable a fair comparison with~\cite{kim2020cross} and~\cite{yue2021prototypical}, we replaced the last fully connected layer with a 512-dimension randomly initialized linear layer. 
% L2-normalizing is performed on the output features. 
We use $k$-means GPU implementation in faiss~\cite{faiss} for efficient clustering. We use SGD with a momentum of 0.9, a learning rate of 0.01, a batch size of 64. More implementation details can be found in the appendix.

\subsection{Results on MFDA}
\paragraph{Baselines.}\label{para:baseline} We compare MSFAN with the following methods. \textbf{(1) Source-only}, \textit{i.e.} train on the labeled data in source domains and test on the target domain directly. \textbf{(2) Single-source DA}, perform multi-source DA via single-source DA, including CDAN~\cite{long2018conditional}, MDDIA~\cite{jiang2020implicit}, MME~\cite{saito2019semi}; as well as CDS~\cite{kim2020cross} and PCS~\cite{yue2021prototypical} which are the strongest single-source baselines specifically designed for single-source few-shot DA (FUDA). 
\textbf{(3) Multi-source DA}, assume multiple fully-labeled sources and are designed for MDA, including MFSAN~\cite{zhu2019aligning}, SImpAl~\cite{venkat2021your} and ProtoMDA~\cite{zhou2021prototype}. SImpAl~\cite{venkat2021your} and ProtoMDA~\cite{zhou2021prototype} are the most recent state-of-the-art works, and ProtoMDA also leverages prototypes for MDA. We re-run all baseline methods in the new MFDA setting (multi-source domain adaptation with few labels in each source), and compare with the proposed MSFAN.

\begin{table}[]
\centering
\caption{Adaptation accuracy (\%) with 1 and 3 labeled samples per class on Office dataset. }
\resizebox{0.97\textwidth}{!}{%
\begin{tabular}{@{}clcccc|cccc@{}}
\toprule
\multicolumn{10}{c}{Office} \\ \midrule
\multicolumn{1}{l}{\multirow{2}{*}{\textbf{}}} & \textbf{} & \multicolumn{4}{c}{1-shot} & \multicolumn{4}{c}{3-shot} \\ \cmidrule(l){2-10} 
% \multicolumn{1}{l}{} & \multicolumn{1}{c|}{Method} & \begin{tabular}[c]{@{}c@{}}D,W\\ $\rightarrow$A\end{tabular} & \begin{tabular}[c]{@{}c@{}}A,W\\ $\rightarrow$D\end{tabular} & \begin{tabular}[c]{@{}c@{}}A,D\\ $\rightarrow$W\end{tabular} & \multicolumn{1}{c|}{Avg} & \begin{tabular}[c]{@{}c@{}}D,W\\ $\rightarrow$A\end{tabular} & \begin{tabular}[c]{@{}c@{}}A,W\\ $\rightarrow$D\end{tabular} & \begin{tabular}[c]{@{}c@{}}A,D\\ $\rightarrow$W\end{tabular} & Avg \\ \midrule
\multicolumn{1}{l}{} & Method & D,W$\rightarrow$A & A,W$\rightarrow$D & A,D$\rightarrow$W & Avg & D,W$\rightarrow$A & A,W$\rightarrow$D & A,D$\rightarrow$W & \multicolumn{1}{c}{Avg} \\ \midrule
\multicolumn{1}{c|}{\multirow{2}{*}{Source Only}} & \multicolumn{1}{l|}{Single-best} & 41.1 & 62.0 & 65.2 & \multicolumn{1}{c|}{56.1} & 55.3 & 86.1 & 85.5 & \multicolumn{1}{c}{75.6} \\
\multicolumn{1}{c|}{} & \multicolumn{1}{l|}{Combined} & 53.4 & 66.5 & 69.2 & \multicolumn{1}{c|}{63.0} & 63.5 & 86.9 & 86.0 & \multicolumn{1}{c}{78.8} \\ \midrule
\multicolumn{1}{c|}{\multirow{5}{*}{\begin{tabular}[c]{@{}c@{}}Single-best\\ DA\end{tabular}}} & \multicolumn{1}{l|}{CDAN~\cite{long2018conditional}} & 39.7 & 66.8 & 66.5 & \multicolumn{1}{c|}{57.7} & 65.1 & 89.8 & 91.6 & \multicolumn{1}{c}{82.2} \\
\multicolumn{1}{c|}{} & \multicolumn{1}{l|}{MME~\cite{saito2019semi}} & 23.1 & 62.4 & 60.9 & \multicolumn{1}{c|}{48.8} & 60.2 & 91.4 & 89.7 & \multicolumn{1}{c}{80.4} \\
\multicolumn{1}{c|}{} & \multicolumn{1}{l|}{MDDIA~\cite{jiang2020implicit}} & 55.6 & 79.5 & 84.4 & \multicolumn{1}{c|}{73.2} & 70.3 & 93.2 & 93.3 & \multicolumn{1}{c}{85.6} \\
\multicolumn{1}{c|}{} & \multicolumn{1}{l|}{CDS~\cite{kim2020cross}} & 52.0 & 57.4 & 59.0 & \multicolumn{1}{c|}{56.1} & 67.6 & 81.3 & 86.0 & \multicolumn{1}{c}{78.3} \\
\multicolumn{1}{c|}{} & \multicolumn{1}{l|}{PCS~\cite{yue2021prototypical}} & 76.1 & 91.8 & 90.6 & \multicolumn{1}{c|}{86.2} & 76.4 & \textbf{96.0} & 94.1 & \multicolumn{1}{c}{88.8} \\ \midrule
\multicolumn{1}{c|}{\multirow{5}{*}{\begin{tabular}[c]{@{}c@{}}Source-combined\\ DA\end{tabular}}} & \multicolumn{1}{l|}{CDAN~\cite{long2018conditional}} & 52.3 & 72.7 & 73.3 & \multicolumn{1}{c|}{66.1} & 67.8 & 85.7 & 88.5 & \multicolumn{1}{c}{80.7} \\
\multicolumn{1}{c|}{} & \multicolumn{1}{l|}{MME~\cite{saito2019semi}} & 34.6 & 64.9 & 74.1 & \multicolumn{1}{c|}{57.9} & 61.5 & 91.2 & 91.4 & \multicolumn{1}{c}{81.4} \\
\multicolumn{1}{c|}{} & \multicolumn{1}{l|}{MDDIA~\cite{jiang2020implicit}} & 63.4 & 91.4 & 87.2 & \multicolumn{1}{c|}{80.7} & 74.7 & 96.6 & 94.9 & \multicolumn{1}{c}{88.7} \\
\multicolumn{1}{c|}{} & \multicolumn{1}{l|}{CDS~\cite{kim2020cross}} & 67.1 & 73.9 & 88.2 & \multicolumn{1}{c|}{76.4} & 72.2 & 88.2 & 90.9 & \multicolumn{1}{c}{83.8} \\
\multicolumn{1}{c|}{} & \multicolumn{1}{l|}{PCS~\cite{yue2021prototypical}} & 72.8 & 89.0 & 92.1 & \multicolumn{1}{c|}{84.6} & 76.5 & \textbf{96.0} & 94.8 & \multicolumn{1}{c}{89.1} \\ \midrule
\multicolumn{1}{c|}{\multirow{4}{*}{\begin{tabular}[c]{@{}c@{}}Multi-source\\ DA\end{tabular}}} & \multicolumn{1}{l|}{SImpAl~\cite{venkat2021your}} & 58.5 & 72.5 & 71.7 & \multicolumn{1}{c|}{67.6} & 65.0 & 85.3 & 86.7 & \multicolumn{1}{c}{79.0} \\
\multicolumn{1}{c|}{} & \multicolumn{1}{l|}{MFSAN~\cite{zhu2019aligning}} &48.9 & 64.7 & 66.0 & 59.9& 64.7 & 82.7 & 87.9 & \multicolumn{1}{c}{78.4} \\
\multicolumn{1}{c|}{} & \multicolumn{1}{l|}{PMDA~\cite{zhou2021prototype}} & 56.3 & 66.5 & 71.4 & \multicolumn{1}{c|}{64.7} & 68.4 & 86.5 & 91.8 & \multicolumn{1}{c}{82.2} \\ 
%\cmidrule(l){2-10} 
\multicolumn{1}{c|}{} & \multicolumn{1}{l|}{\Gc MSFAN (Ours)} & \Gc\textbf{76.3} & \Gc\textbf{94.4} & \Gc\textbf{92.6} & \multicolumn{1}{c|}{\Gc\textbf{87.8}} & \Gc\textbf{77.7} & \Gc95.4 & \Gc\textbf{95.8} & \multicolumn{1}{c}{\Gc\textbf{89.6}} \\ \bottomrule
\end{tabular}
}
\label{tab:office}
\end{table}

\begin{table}[]
\centering
\caption{Adaptation accuracy (\%) with 3\% and 6\% labeled samples per class on Office-Home dataset. }
\resizebox{1.0\textwidth}{!}{%
\begin{tabular}{@{}cccccccccccl@{}}
\toprule
\multicolumn{12}{c}{Office-Home} \\ \midrule
\multicolumn{1}{l}{\textbf{}} & \multicolumn{1}{l}{\textbf{}} & \multicolumn{5}{c}{3\%} & \multicolumn{5}{c}{6\%} \\ \cmidrule(l){2-12} 
\textbf{} & \multicolumn{1}{c|}{Method} & \begin{tabular}[c]{@{}c@{}}Ar,Pr,Rw\\ $\rightarrow$Cl\end{tabular} & \begin{tabular}[c]{@{}c@{}}Cl,Pr,Rw\\ $\rightarrow$Ar\end{tabular} & \begin{tabular}[c]{@{}c@{}}Cl,Ar,Rw\\ $\rightarrow$Pr\end{tabular} & \begin{tabular}[c]{@{}c@{}}Cl,Ar,Pr\\ $\rightarrow$Rw\end{tabular} & \multicolumn{1}{c|}{Avg} & \begin{tabular}[c]{@{}c@{}}Ar,Pr,Rw\\ $\rightarrow$Cl\end{tabular} & \begin{tabular}[c]{@{}c@{}}Cl,Pr,Rw\\ $\rightarrow$Ar\end{tabular} & \begin{tabular}[c]{@{}c@{}}Cl,Ar,Rw\\ $\rightarrow$Pr\end{tabular} & \begin{tabular}[c]{@{}c@{}}Cl,Ar,Pr\\ $\rightarrow$Rw\end{tabular} & \multicolumn{1}{c}{Avg} \\ \midrule
\multicolumn{1}{c|}{\multirow{2}{*}{Source Only}} & \multicolumn{1}{l|}{Single-best} & 29.0 & 41.2 & 52.3 & 43.1 & \multicolumn{1}{c|}{41.4} & 36.0 & 49.9 & 61.8 & 54.6 & 50.6 \\
\multicolumn{1}{c|}{} & \multicolumn{1}{l|}{Combined} & 42.2 & 55.3 & 63.6 & 64.1 & \multicolumn{1}{c|}{56.3} & 45.3 & 60.4 & 70.5 & 70.9 & 61.8 \\ \midrule
\multicolumn{1}{c|}{\multirow{5}{*}{\begin{tabular}[c]{@{}c@{}}Single-best\\ DA\end{tabular}}} & \multicolumn{1}{l|}{CDAN~\cite{long2018conditional}} & 27.0 & 38.7 & 44.9 & 40.3 & \multicolumn{1}{c|}{37.7} & 40.1 & 54.9 & 63.6 & 59.3 & 54.5 \\
\multicolumn{1}{c|}{} & \multicolumn{1}{l|}{MME~\cite{saito2019semi}} & 29.0 & 39.3 & 52.0 & 44.9 & \multicolumn{1}{c|}{41.3} & 37.3 & 54.9 & 66.8 & 61.3 & 55.1 \\
\multicolumn{1}{c|}{} & \multicolumn{1}{l|}{MDDIA~\cite{jiang2020implicit}} & 29.5 & 47.1 & 56.4 & 51.0 & \multicolumn{1}{c|}{46.0} & 37.1 & 58.2 & 68.4 & 64.5 & 57.1 \\
\multicolumn{1}{c|}{} & \multicolumn{1}{l|}{CDS~\cite{kim2020cross}} & 37.8 & 51.6 & 53.8 & 51.0 & \multicolumn{1}{c|}{48.6} & 45.3 & 63.7 & 68.6 & 65.2 & 60.7 \\
\multicolumn{1}{c|}{} & \multicolumn{1}{l|}{PCS~\cite{yue2021prototypical}} & 52.5 & 66.0 & \textbf{75.6} & 73.9 & \multicolumn{1}{c|}{67.0} & 54.7 & 67.0 & 76.6 & 75.2 & 68.4 \\ \midrule
\multicolumn{1}{c|}{\multirow{5}{*}{\begin{tabular}[c]{@{}c@{}}Source-combined\\ DA\end{tabular}}} & \multicolumn{1}{l|}{CDAN~\cite{long2018conditional}} & 42.6 & 52.3 & 64.5 & 63.2 & \multicolumn{1}{c|}{55.7} & 51.1 & 67.0 & 74.2 & 73.3 & 66.4 \\
\multicolumn{1}{c|}{} & \multicolumn{1}{l|}{MME~\cite{saito2019semi}} & 42.5 & 55.4 & 67.4 & 64.5 & \multicolumn{1}{c|}{57.5} & 46.0 & 67.1 & 75.5 & 75.7 & 66.1 \\
\multicolumn{1}{c|}{} & \multicolumn{1}{l|}{MDDIA~\cite{jiang2020implicit}} & 55.3 & 66.9 & 72.3 & 75.3 & \multicolumn{1}{c|}{67.5} & \textbf{57.3} & 67.2 &  79.0 & 74.4 & 69.5 \\
\multicolumn{1}{c|}{} & \multicolumn{1}{l|}{CDS~\cite{kim2020cross}} & 54.9 & 66.2 & 71.6 & 73.4 & \multicolumn{1}{c|}{66.5} & 54.9 & 67.5 & 76.1 & 77.5 & 69.0 \\
\multicolumn{1}{c|}{} & \multicolumn{1}{l|}{PCS~\cite{yue2021prototypical}} & 49.4 & 67.0 & 75.0 & 76.3 & \multicolumn{1}{c|}{66.9} & 50.4 & 67.0 & 77.8 & \textbf{79.4} & 68.7 \\ \midrule
\multicolumn{1}{c|}{\multirow{4}{*}{\begin{tabular}[c]{@{}c@{}}Multi-source\\ DA\end{tabular}}} & \multicolumn{1}{l|}{SImpAl~\cite{venkat2021your}} & 46.8 & 56.7 & 65.1 & 66.6 & \multicolumn{1}{c|}{58.8} & 49.3 & 62.1 & 71.7 & 73.0 & 64.1 \\
\multicolumn{1}{c|}{} & \multicolumn{1}{l|}{MFSAN~\cite{zhu2019aligning}} & 39.9 & 46.6 & 58.9 & 55.6 & \multicolumn{1}{c|}{50.3} & 44.5 & 53.7 & 65.4 & 64.2 & \multicolumn{1}{c}{57.0} \\
\multicolumn{1}{c|}{} & \multicolumn{1}{l|}{PMDA~\cite{zhou2021prototype}} & 50.8 & 56.8 & 64.2 & 66.8 & \multicolumn{1}{c|}{59.7} & 54.4 & 65.8 & 70.4 & 71.8 & \multicolumn{1}{c}{65.6} \\ 
%\cmidrule(l){2-12} 
\multicolumn{1}{c|}{} & \multicolumn{1}{l|}{\Gc MSFAN (Ours)} & \Gc\textbf{55.6} & \Gc\textbf{68.4} & \Gc\textbf{75.6} & \Gc\textbf{76.6} & \multicolumn{1}{c|}{\Gc\textbf{69.1}} & \Gc56.3 & \Gc\textbf{68.7} & \Gc\textbf{79.3} & \Gc79.1 & \multicolumn{1}{c}{\Gc\textbf{70.9}} \\ \bottomrule
\end{tabular}
}
\label{tab:officehome}
\end{table}

Extensive experiments are performed on Office, Office-Home, and DomainNet, with the results shown in Table~\ref{tab:office}, \ref{tab:officehome}, \ref{tab:domainnet}, respectively. 
From the results, we have the following observations: (1) For single-best, source-only outperforms some UDA methods under various scenarios, \textit{e.g.} 56.1\% vs. 48.8\% in Office under 1-shot per class. A similar observation is obtained on source-combined, \textit{e.g.} 40.1\% vs. 31.3\% in Office-Home under 1-shot per class. (2) Under MFDA, naively combining multiple sources and perform single-source DA can lead to worse performance than using a single domain, \textit{e.g.} 42.3\% vs. 44.6\% with 1-shot per class on DomainNet for PCS, which is specifically designed for single-source few-shot DA. (3) Under MFDA, conventional MDA methods perform even worse than single-source DA methods, \textit{e.g.} 65.6\% vs. 68.7\% with 6\% labels per class on Office-Home. (4) MSFAN outperforms all baselines under all experimental settings. Especially, compared to state-of-the-art MDA methods, we can see that MSFAN outperforms them across all benchmarks with large improvements: 20.2\% and 7.4\% on Office, 9.4\% and 5.3\% on Office-Home, 16.2\% and 9.0\% on DomainNet. 

\begin{table}[]
\centering
\caption{Adaptation accuracy (\%) comparison with 1 and 3 labeled samples per class on DomainNet. }
\resizebox{1.0\textwidth}{!}{%
\begin{tabular}{@{}cccccccccccc@{}}
\toprule
\multicolumn{12}{c}{DomainNet} \\ \midrule
\multicolumn{1}{l}{\textbf{}} & \multicolumn{1}{l}{\textbf{}} & \multicolumn{5}{c}{1-shot} & \multicolumn{5}{c}{3-shot} \\ \cmidrule(l){2-12} 
\multicolumn{1}{l}{\textbf{}} & \multicolumn{1}{c|}{Method} & \begin{tabular}[c]{@{}c@{}}P,R,S\\ $\rightarrow$C\end{tabular} & \begin{tabular}[c]{@{}c@{}}C,R,S\\ $\rightarrow$P\end{tabular} & \begin{tabular}[c]{@{}c@{}}C,P,S\\ $\rightarrow$R\end{tabular} & \begin{tabular}[c]{@{}c@{}}C,P,R\\ $\rightarrow$S\end{tabular} & \multicolumn{1}{c|}{Avg} & \begin{tabular}[c]{@{}c@{}}P,R,S\\ $\rightarrow$C\end{tabular} & \begin{tabular}[c]{@{}c@{}}C,R,S\\ $\rightarrow$P\end{tabular} & \begin{tabular}[c]{@{}c@{}}C,P,S\\ $\rightarrow$R\end{tabular} & \begin{tabular}[c]{@{}c@{}}C,P,R\\ $\rightarrow$S\end{tabular} & \multicolumn{1}{c}{Avg} \\ \midrule
\multicolumn{1}{c|}{\multirow{2}{*}{Source Only}} & \multicolumn{1}{l|}{Single Best} & 18.4 & 30.6 & 28.9 & 16.7 & \multicolumn{1}{c|}{23.7} & 30.2 & 44.2 & 49.8 & 24.2 & \multicolumn{1}{c}{ 34.4} \\
\multicolumn{1}{c|}{} & \multicolumn{1}{l|}{Combined} & 30.8 & 49.4 & 43.3 & 36.9 & \multicolumn{1}{c|}{40.1} & 45.3 & 57.4 & 64.7 & 42.6 & \multicolumn{1}{c}{ 50.0} \\ \midrule
\multicolumn{1}{c|}{\multirow{5}{*}{\begin{tabular}[c]{@{}c@{}}Single-best\\ DA\end{tabular}}} & \multicolumn{1}{l|}{CDAN~\cite{long2018conditional}} & 16.0 & 25.7 & 19.5 & 12.9 & \multicolumn{1}{c|}{18.5} & 30.0 & 40.1 & 40.8 & 17.1 & \multicolumn{1}{c}{ 29.3} \\
\multicolumn{1}{c|}{} & \multicolumn{1}{l|}{MME~\cite{saito2019semi}} & 16.0 & 29.2 & 26.0 & 13.4 & \multicolumn{1}{c|}{21.2} & 25.1 & 46.5 & 50.0 & 20.1 & \multicolumn{1}{c}{ 32.6} \\
\multicolumn{1}{c|}{} & \multicolumn{1}{l|}{MDDIA~\cite{jiang2020implicit}} & 18.0 & 30.6 & 27.4 & 15.9 & \multicolumn{1}{c|}{23.0} & 41.4 & 50.7 & 52.9 & 23.1 & \multicolumn{1}{c}{ 38.2} \\
\multicolumn{1}{c|}{} & \multicolumn{1}{l|}{CDS~\cite{kim2020cross}} & 16.7 & 24.4 & 15.9 & 13.4 & \multicolumn{1}{c|}{17.6} & 35.0 & 43.8 & 36.8 & 31.1 & \multicolumn{1}{c}{ 32.9} \\
\multicolumn{1}{c|}{} & \multicolumn{1}{l|}{PCS~\cite{yue2021prototypical}} & 39.0 & 51.7 & 38.8 & 39.8 & \multicolumn{1}{c|}{42.3} & 45.2 & 59.1 & 66.6 & 41.9 & \multicolumn{1}{c}{ 51.0} \\ \midrule
\multicolumn{1}{c|}{\multirow{5}{*}{\begin{tabular}[c]{@{}c@{}}Source-combined\\ DA\end{tabular}}} & \multicolumn{1}{l|}{CDAN~\cite{long2018conditional}} & 25.7 & 33.0 & 40.0 & 26.4 & \multicolumn{1}{c|}{31.3} & 47.8 & 54.1 & 65.6 & 49.1 & \multicolumn{1}{c}{ 49.6} \\
\multicolumn{1}{c|}{} & \multicolumn{1}{l|}{MME~\cite{saito2019semi}} & 20.0 & 45.3 & 52.5 & 13.0 & \multicolumn{1}{c|}{32.7} & 44.2 & 62.7 & 73.9 & 51.8 & \multicolumn{1}{c}{ 53.1} \\
\multicolumn{1}{c|}{} & \multicolumn{1}{l|}{MDDIA~\cite{jiang2020implicit}} & 44.0 & 46.4 & 49.6 & 37.1 & \multicolumn{1}{c|}{44.3} & 56.3 & 59.3 & 70.3 & 51.3 & \multicolumn{1}{c}{ 56.3} \\
\multicolumn{1}{c|}{} & \multicolumn{1}{l|}{CDS~\cite{kim2020cross}} & 42.2 & 53.3 & 55.4 & 38.5 & \multicolumn{1}{c|}{47.4} & 50.2 & 61.5 & 71.8 & 47.3 & \multicolumn{1}{c}{ 55.6} \\
\multicolumn{1}{c|}{} & \multicolumn{1}{l|}{PCS~\cite{yue2021prototypical}} & 36.2 & 53.0 & 56.4 & 32.8 & \multicolumn{1}{c|}{44.6} & 45.6 & 61.2 & 74.3 & 41.3 & \multicolumn{1}{c}{ 53.4} \\ \midrule
\multicolumn{1}{c|}{\multirow{4.5}{*}{\begin{tabular}[c]{@{}c@{}}Multi-source\\ DA\end{tabular}}} & \multicolumn{1}{l|}{SImpAl~\cite{venkat2021your}} & 48.0 & 40.3 & 45.7 & 35.3 & \multicolumn{1}{c|}{42.3} & 51.5 & 47.4 & 68.8 & 45.3 & \multicolumn{1}{c}{ 51.1} \\
\multicolumn{1}{c|}{} & \multicolumn{1}{l|}{MFSAN~\cite{zhu2019aligning}} & 41.6 & 33.5 & 38.8 & 29.6 & \multicolumn{1}{c|}{35.9} & 43.5 & 42.3 & 63.2 & 41.1 & \multicolumn{1}{c}{ 45.2} \\ 
\multicolumn{1}{c|}{} & \multicolumn{1}{l|}{PMDA~\cite{zhou2021prototype}} & 49.3 & 42.2 & 45.0 & 34.8 & \multicolumn{1}{c|}{42.8} & 52.2 & 52.5 & 71.3 & 47.6 & \multicolumn{1}{c}{ 53.3} \\ %\cmidrule(l){2-12} 
\multicolumn{1}{c|}{} & \multicolumn{1}{l|}{\Gc MSFAN (Ours)} & \Gc\textbf{57.3} & \Gc\textbf{68.7} & \Gc\textbf{64.8} & \Gc\textbf{45.2} & \multicolumn{1}{c|}{\Gc\textbf{59.0}} & \Gc\textbf{57.8} & \Gc\textbf{65.5} & \Gc\textbf{75.8} & \Gc\textbf{53.6} & \multicolumn{1}{c}{\Gc \textbf{62.3}} \\ \bottomrule
\end{tabular}
}
\label{tab:domainnet}
\end{table}
% 49.3	42.2	45.0	34.8	42.8	52.2	52.5	71.3	47.6	53.3
\begin{table}[]
\centering
\caption{Performance contribution of each part in MSFAN framework on Office-Home.}
\resizebox{0.7\textwidth}{!}{%
\begin{tabular}{@{}lccccc@{}}
\toprule
\multicolumn{1}{c|}{Office-Home 3\%} & \begin{tabular}[c]{@{}c@{}}Ar,Pr,Rw\\ $\rightarrow$Cl\end{tabular} & \begin{tabular}[c]{@{}c@{}}Cl,Pr,Rw\\ $\rightarrow$Ar\end{tabular} & \begin{tabular}[c]{@{}c@{}}Cl,Ar,Rw\\ $\rightarrow$Pr\end{tabular} & \begin{tabular}[c]{@{}c@{}}Cl,Ar,Pr\\ $\rightarrow$Rw\end{tabular} & \multicolumn{1}{c}{Avg} \\ \midrule
% Officehome 3\% & Ar,Pr,Rw $\rightarrow$Cl & Cl,Pr,Rw $\rightarrow$Ar & Cl,Ar,Rw $\rightarrow$Pr & Cl,Ar,Pr $\rightarrow$Rw\\
\multicolumn{1}{l|}{Source-combined} & 42.2 & 55.3 & 63.6 & 64.1 & \multicolumn{1}{c}{56.3} \\
\multicolumn{1}{l|}{+ Multi-classifier} & 44.9 & 55.5 & 65.1 & 68.2 & \multicolumn{1}{c}{58.4} \\
\multicolumn{1}{l|}{+ $\mathcal{L}_\mathrm{MPS}$} & 52.1 & 66.6 & 66.5 & 75.1 & \multicolumn{1}{c}{65.1} \\
\multicolumn{1}{l|}{+ $\mathcal{L}_\mathrm{MI}$} & 55.2 & \textbf{68.4} & 75.0 & 76.4 & \multicolumn{1}{c}{68.8} \\
\Gd \multicolumn{1}{l|}{+ $\mathcal{L}_\mathrm{SSC}$ (MSFAN)} & \textbf{55.6} & \textbf{68.4} & \textbf{75.6} & \textbf{76.6} & \multicolumn{1}{c}{\Gc\textbf{69.1}} \\ \bottomrule
\end{tabular}
}
\label{tab:officehome-ablation}
\end{table}

\begin{figure*}[t]
 \centering
 \includegraphics[width=\textwidth, trim={0.0cm 0cm 0cm 0cm}]{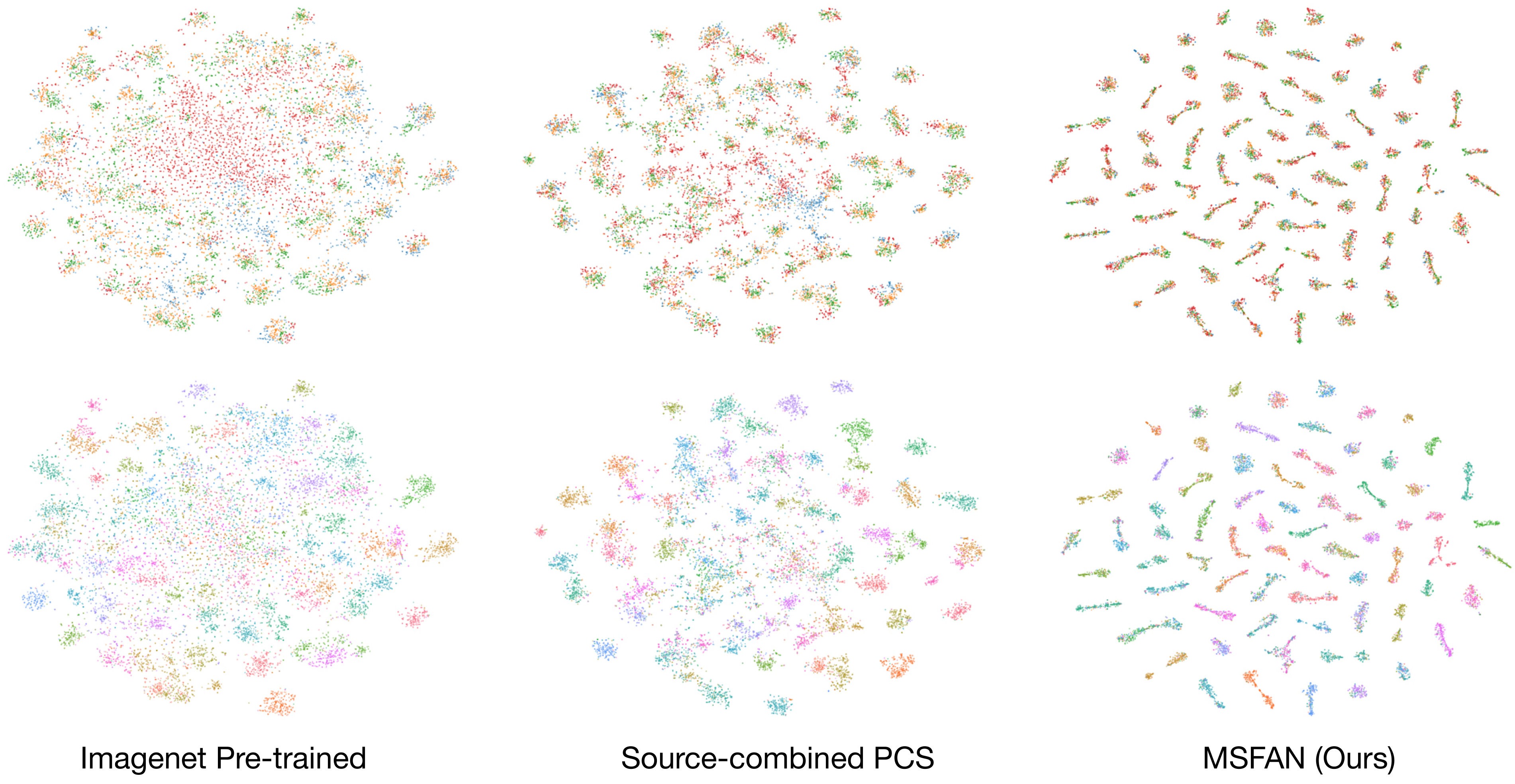}
 %\fbox{\rule[-.5cm]{0cm}{4cm} \rule[-.5cm]{4cm}{0cm}}
 \caption{Visualization of baselines and our method via t-SNE on OfficeHome (Ar,Pr,Rw$\rightarrow$Cl). Top row: \textcolor{tsneBlue}{Blue}, \textcolor{tsneOrange}{Orange}, \textcolor{tsneGreen}{Green} and \textcolor{tsneRed}{Red} represents domain Art, Real, Product and Clipart, respectively. Bottom row: Coloring represents the class of each sample. Features from MSFAN are better-aligned across domains compared to other methods.
%  between domains compared to other methods.
 }
 \label{fig:tsne-officehome}
% \vspace{-2mm}
\end{figure*}

% \textbf{CDAN}~\cite{long2018conditional} and \textbf{MDDIA}~\cite{jiang2020implicit} are both state-of-the-art methods in UDA with a domain classifier to perform domain alignment. \textbf{MME}~\cite{saito2019semi} minimizes the conditional entropy of unlabeled target data with respect to the feature extractor and maximizes it with respect to the classifier. \textbf{CAN}~\cite{kang2019contrastive} uses clustering information to contrast discrepancy of source and target domain. \textbf{CDS}~\cite{kim2020cross} is a instance-based cross-domain self-supervised pre-training, which can be used for other domain adaptation methods and form \textit{two-stage} methods, such as \textbf{CDS / CDAN} and \textbf{CDS / MME}. We re-implement CDS into an end-to-end version by adding losses in two stage together and tuning the weight for different losses. We also investigate the \textit{one-stage} version of the methods above (\textbf{CDS + CDAN}, \textbf{CDS + MME}). Following~\cite{kim2020cross}, entropy minimization (\textbf{ENT}) on source is added to previous DA methods to obtain better baseline performance. 

\subsection{Ablation Study and Analysis}
We now investigate the effectiveness of each component in MSFAN on Office-Home. Table~\ref{tab:officehome-ablation} shows that adding each component contributes to the final MFDA performance without any accuracy degradation. To qualitatively show the effectiveness of domain alignment with MSFAN, we plot the learned features with t-SNE~\cite{maaten2008visualizing} on the Ar,Cl,Pr$\rightarrow$Rl setting in Office-Home.
In the top row, color represents the domain of each sample; while in the bottom row, color represents the class of each sample. Compared to ImageNet pre-training and Source-combined, it qualitatively shows that MSFAN clusters samples with the same class in the feature space; thus, MSFAN favors more discriminative features. Also, the features from MSFAN are more closely aggregated than ImageNet pre-training and Source-combined, which demonstrates that MSFAN learns a better semantic structure of the datasets.

\section{Related Work and Discussion}
\paragraph{Single-source UDA}
Single-source UDA~\cite{gopalan2011domain} aims to transfer knowledge from a fully-labeled source domain to an unlabeled target domain. Most UDA methods focus on feature distribution alignment. Discrepancy-based methods
utilize different metric learning schemas to diminish the domain shift between source and target. Inspired by the two-sample test~\cite{gretton2012kernel}, Maximum Mean Discrepancy (MMD) is leveraged to perform domain alignment in various methods~\cite{long2015learning, tzeng2014deep, long2016unsupervised,ghifary2014domain,wang2018visual,long2017deep}. Sun \etal~\cite{sun2017correlation} and Zhuo \etal~\cite{zhuo2017deep} further proposed to align second-order statistics of source and target features. 
After Generative Adversarial Network~\cite{goodfellow2014generative} was proposed, more works~\cite{ganin2015unsupervised, tzeng2017adversarial, hoffman2018cycada, xie2018learning, long2018conditional, shen2017wasserstein} leverage a domain discriminator to encourage domain confusion by an adversarial objective. Recently, image translation methods~\cite{zhu2017unpaired,liu2016coupled} have been adopted to further improve domain adaptation by performing domain alignment at pixel-level~\cite{hoffman2018cycada, bousmalis2017unsupervised, russo2018source, murez2018image, yue2019domain, sankaranarayanan2018generate, shrivastava2017learning}. Instead of explicit feature alignment, Saito \etal~\cite{saito2019semi} perform entropy optimization for adaptation. Though these methods achieves high performance, few of them consider the practical scenario of adapting from multiple sources. 

\paragraph{Multi-source Domain Adaptation (MDA).} MDA~\cite{sun2015survey, zhao2019multi} assumes the availability of multiple fully-labeled sources and aims to transfer knowledge to an unlabeled target domain. Various theoretical analyses~\cite{ben2010theory, crammer2008learning,mansour2008domain, hoffman2018algorithms} have been proposed to support existing MDA algorithms. Early MDA methods usually either learn a shared feature space for all domains~\cite{duan2009domain,sun2011two,duan2012exploiting,duan2012domain}, or combine pre-learned source classifier predictions to get final predictions with an ensembling method. With the development of deep nerual networks, more deep-learning-based MDA methods are proposed, such as DCTN~\cite{xu2018deep}, M3SDA~\cite{peng2019moment}, MDAN~\cite{zhao2018adversarial}, MFSAN~\cite{zhu2019aligning}, MDDA~\cite{zhao2020multi}. All these MDA methods aim to minimize this domain shift using auxiliary
distribution alignment objectives. SImpAl~\cite{venkat2021your} is proposed to perform implicit domain alignment with pseudo-labeling without additional training objectives for adaptation. Recently, ProtoMDA~\cite{zhou2021prototype} is proposed to use prototypes for MDA and achieves state-of-the-art performance. While these methods have full supervision on the source domains, we focus on a new adaptation setting with only few labels in each source domain.
% employ a shared feature extractor network to symmetrically map the multiple sources and target into the same space. 
% Existing methods aim to minimize this domain-shift using auxiliary
% distribution alignment objectives. In this work, we present a different perspective
% to MSDA wherein deep models are observed to implicitly align the domains under
% label supervision.
\paragraph{Self-supervised Learning (SSL) for Domain Adaptation.} SSL is a subset of unsupervised learning where supervision is automatically generated from the data~\cite{jing2020self, doersch2015unsupervised, zhang2016colorful,noroozi2016unsupervised,gidaris2018unsupervised,wang2020pre}. One of the most common strategies for SSL is handcrafting auxiliary pretext tasks predicting future, missing or contextual information~\cite{zhang2016colorful, larsson2016learning,doersch2015unsupervised, doersch2017multi, noroozi2016unsupervised,pathak2016context,gidaris2018unsupervised,dosovitskiy2014discriminative}. 
Reconstruction was first utilized as a self-supervised task in some early works~\cite{ghifary2015domain, ghifary2016deep}, in which source and target share the same encoder to extract domain-invariant features. 
% To capture both domain-specific and shared properties, Bousmalis \etal~\cite{bousmalis2017unsupervised} explicitly extracts image representations into two spaces, one private for each domain and one shared across domains.
In~\cite{carlucci2019domain}, solving jigsaw puzzle~\cite{noroozi2016unsupervised} was leveraged as a self-supervision task to solve domain adaptation and generalization. Sun \etal~\cite{sun2019unsupervised} further proposed to perform adaptation by jointly learning multiple self-supervision tasks. 
% The feature encoder is shared by both source and target images, and the extracted features are then fed into different self-supervision task heads.
% Doersch and Zisserman~\cite{doersch2017multi} further investigated combining multiple self-supervised tasks for better representation learning. 
Recently, contrastive learning has achieved state-of-the-art performance on representation learning~\cite{he2020momentum,grill2020bootstrap, chen2020simple, chen2020improved, chen2020big, reed2021self, xiao2021region, li2020prototypical, asano2020self, caron2020unsupervised}.
Based on instance discrimination~\cite{wu2018unsupervised} and prototypical contrastive learning, Kim~\etal~\cite{kim2020cross} and Yue~\etal~\cite{yue2021prototypical} proposed cross-domain SSL approaches for adaptation with few source labels. SSL has also been incorporated for adaptation in other fields, including point cloud recognition~\cite{achituve2020self}, medical imaging~\cite{ihler2020self}, action segmentation~\cite{chen2020action}, robotics~\cite{jeong2020self}, facial tracking~\cite{yoon2019self}, \textit{etc}.

\section{Conclusion}
Traditional Multi-source Domain Adaptation assumes multiple fully-labeled source domains. In this paper, we investigate Multi-source Few-shot Domain Adaptation, a new domain adaptation task that is more practical and challenging, where each source domain only has a very small fraction of labeled samples. 
We proposed a novel framework, termed Multi-Source Few-shot Adaptation Network (MSFAN),  that performs multi-domain prototypical self-supervised learning, support-set-based cross-domain similarity consistency, and multi-domain prototypical classifier learning. 
We perform extensive experiments on multiple benchmark datasets, which demonstrates the superiority of MSFAN over previous state-of-the-art methods. 
% PCS sets a new state of the art for Few-shot Unsupervised Domain Adaptation. 

\textbf{Acknowledgement.} This work was partially supported by C3.AI DTI, Berkeley AI Research, and by the Berkeley Deep Drive center. We thank Dequan Wang for valuable discussion, and Kostadin Ilov for providing system assistance.

{
\bibliographystyle{unsrt}
\bibliography{ref.bib}
}

\clearpage
%%%%%%%%%%%%%%%%%%%%%%%%%%%%%%%%%%%%%%%%%%%%%%%%%%%%%%%%%%%%
% \section*{Checklist}

%%%%%%%%%%%%%%%%%%%%%%%%%%%%%%%%%%%%%%%%%%%%%%%%%%%%%%%%%%%%
\appendix
\noindent\textbf{\Large Appendix}
\section{Additional Dataset Details}
In Table~\ref{tab:dataset}, we provide more dataset statistics and number of labeled samples in each domain. We follow the dataset setting in~\cite{kim2020cross,yue2021prototypical}. For both Office~\cite{saenko2010adapting} and DomainNet~\cite{peng2019moment}, we use 1-shot and 3-shots labeled samples per class. For Office-Home~\cite{venkateswara2017deep}, we use 3\% and 6\% labeled samples per class. 

\begin{table}[h]
\centering
\caption{Dataset details and labeled sources}
% \vspace{1mm}
\label{tab:dataset}
\resizebox{0.8\textwidth}{!}{
\begin{tabular}{ccccc}
\toprule[1.3pt]
Dataset & \multicolumn{1}{c}{Domain} & \# total image & \# labeled images & \# classes \\ \midrule
\multirow{3}{*}{Office~\cite{saenko2010adapting}} & \multicolumn{1}{c}{Amazon (A)} & 2817 & \multirow{3}{*}{\begin{tabular}[c]{@{}c@{}}1-shot and 3-shots\\ labeled source\end{tabular}} & \multirow{3}{*}{31} \\ \cline{2-3}
 & \multicolumn{1}{c}{DSLR (D)} & 498 &  &  \\ \cline{2-3}
 & \multicolumn{1}{c}{Webcam (W)} & 795 &  &  \\ \midrule
\multirow{4}{*}{Office-Home~\cite{venkateswara2017deep}} & \multicolumn{1}{c}{Art (Ar)} & 2427 & \multirow{4}{*}{\begin{tabular}[c]{@{}c@{}}3\% and 6\%\\ labeled source\end{tabular}} & \multirow{4}{*}{65} \\ \cline{2-3}
 & \multicolumn{1}{c}{Clipart (Cl)} & 4365 &  &  \\ \cline{2-3}
 & \multicolumn{1}{c}{Product (Pr)} & 4439 &  &  \\ \cline{2-3}
 & \multicolumn{1}{c}{Real (Rw)} & 4357 &  &  \\ \midrule
% \multirow{2}{*}{VisDA~\cite{peng2017visda}} & \multicolumn{1}{c}{Synthetic (Syn)} & 152K & \multirow{2}{*}{\begin{tabular}[c]{@{}c@{}}0.1\% and 1\%\\ labeled source\end{tabular}} & \multirow{2}{*}{12} \\ \cline{2-3}
%  & \multicolumn{1}{c}{Real (Rw)} & 55K &  &  \\ \midrule
\multirow{4}{*}{DomainNet~\cite{peng2019moment}} & \multicolumn{1}{c}{Clipart (C)} & 18703 & \multirow{4}{*}{\begin{tabular}[c]{@{}c@{}}1-shot and 3-shots\\ labeled source\end{tabular}} & \multirow{4}{*}{126} \\ \cline{2-3}
 & \multicolumn{1}{c}{Painting (P)} & 31502 &  &  \\ \cline{2-3}
 & \multicolumn{1}{c}{Real (R)} & 70358 &  &  \\ \cline{2-3}
 & \multicolumn{1}{c}{Sketch (S)} & 24582 &  &  \\ \bottomrule
\end{tabular}
}
\end{table}

\section{Additional Implementation Details}

We implemented our model in PyTorch~\cite{paszke2019pytorch}. The training setting is adapted from~\cite{yue2021prototypical}. For Office and Office-Home, the temperature $\phi$ is set adaptively according while for DomainNet, $\phi$ is fixed to 0.1 for more stable training. The margin $m$ is always set to 0.1. We set temperature $\tau$ to be 0.1 in all experiments according to~\cite{yue2021prototypical}. The weights for different loss are $\lambda_{\mathrm{mps}}=1,\lambda_{\mathrm{ssc}}=0.1,\lambda_{\mathrm{mi}}=0.1$.

We use a batch size of 64 and train our model on two NVIDIA P100 GPUs. The setting for clustering is same with~\cite{yue2021prototypical} except that we found more frequent clustering yields better results on DomainNet and generate new prototypes every 200 iterations.

\section{Stability Analysis of MSFAN}

To show the performance stability of MSFAN, we conduct multiple runs with three different random seeds. Table~\ref{tab:multirun} reports the averaged accuracy and standard deviation on the 1-shot and 3-shot labels per class settings of Office. From the variance, we can see that the proposed MSFAN framework is experimentally stable. 

\begin{table}[h]
\centering
\caption{Averaged accuracy (\%) and standard deviation of three runs of 1-shot and 3-shots settings on the Office dataset.}
\resizebox{0.93\textwidth}{!}{%
\begin{tabular}{@{}ccc|ccc@{}}
\toprule
\multicolumn{3}{c}{1-shot} & \multicolumn{3}{c}{3-shot} \\ \midrule
  \multicolumn{1}{c}{D,W$\rightarrow$A} & \multicolumn{1}{c}{A,W$\rightarrow$D} & \multicolumn{1}{c|}{A,D$\rightarrow$W}  & \multicolumn{1}{c}{D,W$\rightarrow$A} & \multicolumn{1}{c}{A,W$\rightarrow$D} & \multicolumn{1}{c}{A,D$\rightarrow$W}  \\ \midrule
 \multicolumn{1}{c}{\Gc $76.3\pm 0.32$} & \multicolumn{1}{c}{\Gc$94.4\pm 1.17$} & \multicolumn{1}{c|}{\Gc$92.6\pm 0.08$} & \multicolumn{1}{c}{\Gc$77.7\pm 0.6$} & \multicolumn{1}{c}{\Gc$95.4\pm 0.06$} & \multicolumn{1}{c}{\Gc$95.8\pm 0.10$}  \\ \bottomrule
\end{tabular}
}
\label{tab:multirun}
\end{table}

\section{Traditional Multi-Source DA with Full Source Labels}
We also apply MSFAN to the traditional MDA setting with full source labels. Table~\ref{tab:office_full_label} shows the performance comparison with state-of-the-art UDA and MDA methods on Office. We can see from the results that the proposed MSFAN framework still outperforms all previous methods with fully labeled source domains. This shows the potential wider application of MSFAN, not only in the 
label-scarce setting, but also in label-abundant setting. We hope to test the potential usage of MSFAN to other DA settings, such as multi-source semi-supervised DA, multi-source partial DA, multi-source open-set DA etc. 
% This could be due to the fact that features learned in MSFAN is more class-discriminative and align better with the target domain. 
% Surpringly, we notice that the source-combined without adaptation achieves 91.6\%, outperforming almost all other UDA and MDA methods. 

\begin{table}[h]
\centering
\caption{Adaptation accuracy (\%) with full labels of source domains on Office dataset. }
\resizebox{0.9\textwidth}{!}{%
\begin{tabular}{@{}clccc|c@{}}
\toprule
\multicolumn{6}{c}{Office full-labeled} \\ \midrule

\multicolumn{1}{l}{} & Method & D,W$\rightarrow$A & A,W$\rightarrow$D & A,D$\rightarrow$W & \multicolumn{1}{c}{Avg} \\ \midrule
\multicolumn{1}{c|}{\multirow{2}{*}{Source Only}} & \multicolumn{1}{l|}{Single-best} & 
62.5 & 99.3 & 96.7 & \multicolumn{1}{c}{86.2}  \\

\multicolumn{1}{c|}{} & \multicolumn{1}{l|}{Combined} &
66.3 & 98.8 & 97.7 & \multicolumn{1}{c}{87.6}  \\ \midrule

\multicolumn{1}{c|}{\multirow{5}{*}{\begin{tabular}[c]{@{}c@{}}Single-best\\ DA\end{tabular}}} & \multicolumn{1}{l|}{CDAN~\cite{long2018conditional}} &
71.0 & 100 & 98.6 & \multicolumn{1}{c}{89.9}  \\

\multicolumn{1}{c|}{} & \multicolumn{1}{l|}{MME~\cite{saito2019semi}} & 
69.2 & 100 & 98.7 & \multicolumn{1}{c}{89.3}  \\

\multicolumn{1}{c|}{} & \multicolumn{1}{l|}{MDDIA~\cite{jiang2020implicit}} &
75.3 & 99.8 & 98.7 & \multicolumn{1}{c}{91.3}  \\

\multicolumn{1}{c|}{} & \multicolumn{1}{l|}{CDS~\cite{kim2020cross}} &
75.9 & 100 & 98.6 & \multicolumn{1}{c}{91.5}  \\

\multicolumn{1}{c|}{} & \multicolumn{1}{l|}{PCS~\cite{yue2021prototypical}} & 
\textbf{77.4} & 99.8 & 97.7 & \multicolumn{1}{c}{91.6}  \\ \midrule

\multicolumn{1}{c|}{\multirow{5}{*}{\begin{tabular}[c]{@{}c@{}}Source-combined\\ DA\end{tabular}}} & \multicolumn{1}{l|}{CDAN~\cite{long2018conditional}} & 
73.0 & 99.4 & 98.8 & \multicolumn{1}{c}{90.4}  \\
 
\multicolumn{1}{c|}{} & \multicolumn{1}{l|}{MME~\cite{saito2019semi}} & 
69.2 & 98.7  & 99.2 & \multicolumn{1}{c}{89.0}  \\
 
\multicolumn{1}{c|}{} & \multicolumn{1}{l|}{MDDIA~\cite{jiang2020implicit}} & 
76.1 & 99.4 & 98.2 & \multicolumn{1}{c}{91.2}  \\

\multicolumn{1}{c|}{} & \multicolumn{1}{l|}{CDS~\cite{kim2020cross}} & 
73.9 & 98.8 & 98.4 & \multicolumn{1}{c}{90.4}  \\

\multicolumn{1}{c|}{} & \multicolumn{1}{l|}{PCS~\cite{yue2021prototypical}} & 
76.3  & 98.3 & 97.1 & \multicolumn{1}{c}{90.6}  \\ \midrule

\multicolumn{1}{c|}{\multirow{4}{*}{\begin{tabular}[c]{@{}c@{}}Multi-source\\ DA\end{tabular}}} & \multicolumn{1}{l|}{SImpAl~\cite{venkat2021your}} &
70.6 & 99.2 & 97.4 & \multicolumn{1}{c}{89.0}  \\

\multicolumn{1}{c|}{} & \multicolumn{1}{l|}{MFSAN~\cite{zhu2019aligning}} &
72.7 & 99.5 & 98.5  & \multicolumn{1}{c}{90.2} \\

\multicolumn{1}{c|}{} & \multicolumn{1}{l|}{PMDA~\cite{zhou2021prototype}} &
75.2 & 99.7 & 98.3 & \multicolumn{1}{c}{91.1}  \\ 

%\cmidrule(l){2-10} 
\multicolumn{1}{c|}{} & \multicolumn{1}{l|}{\Gc MSFAN (Ours)} & \Gc 77.0 & \Gc\textbf{100} & \Gc\textbf{98.9} & \multicolumn{1}{c}{\Gc\textbf{92.0}}\\ \bottomrule
\end{tabular}
}
\label{tab:office_full_label}
\end{table}

\section{Ablation Study on Cross-multi-domain Prototypical SSL Design ($\mathcal{L}_\mathrm{CPS}$)}
For $\mathcal{L}_\mathrm{CPS}$ in ``{Cross-multi-domain Prototypical Self-supervised Learning}'' of
Section 2.1 in the main paper, we propose to only minimize $H(P_{j}^{i\veryshortarrow t})$, without considering $H(P_{j}^{t\veryshortarrow i})$ or the similarity vector entropy between source domains.  

In Table~\ref{tab:office-ablation}, we compare the final performances of different designs of $\mathcal{L}_\mathrm{CPS}$ to validate the proposed $\mathcal{L}_\mathrm{CPS}$. 
All experiments are conducted with $\mathcal{L}_\mathrm{cls} + \mathcal{L}_\mathrm{IPS} + \mathcal{L}_\mathrm{CPS}$, with the same $\mathcal{L}_\mathrm{cls}$ and $\mathcal{L}_\mathrm{IPS}$, and different $\mathcal{L}_\mathrm{CPS}$ for each row: i) ``Every domain pair'' refers to minimizing the entropy of instance-prototype similarity vectors between every domain pair of the $M+1$ domains ($M$ source domains and 1 target domain). ii) ``Tgt$\leftrightarrow$Src'' refers to only minimizing the entropy of similarity vectors between each source domain and the target domain including both $H(P_{j}^{i\veryshortarrow t})$ and $H(P_{j}^{t\veryshortarrow i})$. iii) ``Tgt$\rightarrow$Src'' means only considering the $H(P_{j}^{t\veryshortarrow i})$ between each source and the target. iv) ``Src$\rightarrow$Tgt'' means only considering the $H(P_{j}^{i\veryshortarrow t})$ between each source domain and the target domain. From the results in Table~\ref{tab:office-ablation}, we can see that the proposed design (only minimizing $H(P_{j}^{i\veryshortarrow t})$) achieves the best results.

\begin{table}[h]
\centering
\caption{Performance of different designs of $\mathcal{L}_\mathrm{CPS}$ in Cross-multi-domain Prototypical SSL.}
\resizebox{0.7\textwidth}{!}{%
\begin{tabular}{@{}lcccc@{}}
\toprule
\multicolumn{1}{c|}{Office 1-shot} & D,W$\rightarrow$ A & A,W$\rightarrow$ D & A,D$\rightarrow$ W & \multicolumn{1}{c}{Avg} \\ \midrule
% Officehome 3\% & Ar,Pr,Rw $\rightarrow$Cl & Cl,Pr,Rw $\rightarrow$Ar & Cl,Ar,Rw $\rightarrow$Pr & Cl,Ar,Pr $\rightarrow$Rw\\
\multicolumn{1}{l|}{Every domain pair} & 74.7 & 89.7  & 92.1 & \multicolumn{1}{c}{85.5} \\
\multicolumn{1}{l|}{Tgt$\leftrightarrow$Src} & 75.4 & 89.7 & 91.7 & \multicolumn{1}{c}{85.6} \\
\multicolumn{1}{l|}{Tgt$\rightarrow$Src} & 72.0& 85.3& 91.1& \multicolumn{1}{c}{82.8} \\
\multicolumn{1}{l|}{\Gc Src$\rightarrow$Tgt} & \Gc \textbf{75.6} & \Gc 89.7 & \Gc \textbf{91.8} & \multicolumn{1}{c}{\Gc \textbf{85.7}} \\
\bottomrule
\end{tabular}
}
\label{tab:office-ablation}
\end{table}

%cls + proto-each + loss above

\section{Proof of Equation (14)% on $\mathcal{I}(y; \mathbf{x})$
}

Here we show the proof of Equation (14) in the main paper. We begin with the definition
\begin{align}
    \mathcal{I}(y; \mathbf{x}) &= \iint p(y, \mathbf{x}) \log \frac{p(y, \mathbf{x})}{p(y)p(\mathbf{x})} \dd{y} \dd{\mathbf{x}} \\
    \intertext{Rewriting terms we obtain}
    &= \int p(\mathbf{x}) \dd{\mathbf{x}} \int p(y | \mathbf{x}) \log \frac{p(y | \mathbf{x})}{p(y)} \dd{y} \\
    &= \int p(\mathbf{x}) \dd{\mathbf{x}} \int p(y | \mathbf{x}) \log \frac{p(y | \mathbf{x})}{\int p(\mathbf{x}) p(y | \mathbf{x}) \dd{\mathbf{x}}} \dd{y} \\
    \intertext{Then, rewriting both integrals as expectations we obtain}
    &= \mathbb{E}_\mathbf{x}\left[ \int p(y | \mathbf{x}) \log \frac{p(y | \mathbf{x})}{\mathbb{E}_\mathbf{x}[p(y | \mathbf{x})]} \dd{y} \right] \\
    &= \mathbb{E}_\mathbf{x}\left[ \sum_{i = 1}^L p(y_i | \mathbf{x}) \log \frac{p(y_i | \mathbf{x})}{\mathbb{E}_\mathbf{x}[p(y_i | \mathbf{x})]} \right] \\
    &= \mathbb{E}_\mathbf{x}\left[ \sum_{i = 1}^L p(y_i | \mathbf{x}) \log p(y_i | \mathbf{x}) \right] - \mathbb{E}_\mathbf{x} \left[ \sum_{i = 1}^L p(y_i | \mathbf{x}) \log \mathbb{E}_\mathbf{x}[p(y_i | \mathbf{x})] \right] \\
    &= \mathbb{E}_\mathbf{x}\left[ \sum_{i = 1}^L p(y_i | \mathbf{x}) \log p(y_i | \mathbf{x}) \right] -  \sum_{i = 1}^L \mathbb{E}_\mathbf{x}[ p(y_i | \mathbf{x}) ] \log \mathbb{E}_\mathbf{x}[ p(y_i | \mathbf{x})] ] \\
    &= \ent( \mathbb{E}_\mathbf{x} [p(y | \mathbf{x}; \theta)]) - \mathbb{E}_\mathbf{x}[ \ent( p(y | \mathbf{x}; \theta))]
\end{align}
In addition, we estimate $\mathcal{H}(\mathbb{E}_{\mathbf{x}}[p(y|\mathbf{x}; \theta)])$ with $\sum_{\mathbf{x}\in \mathcal{D}} p(y|\mathbf{x}; \theta)\log \mathbf{\hat p}_0$, where $\mathbf{\hat p}_0$ is a moving average of $p(y|\mathbf{x}; \theta)$; and $\mathcal{D} = \bigcup_i\mathcal{S}^u_i \cup \mathcal{T}$. 
% Limitation

\section{Potential Limitation}
One limitation of the work is that it only considers adapting to a fixed target domain. In the future, we plan to extend it to Domain Generalization setting, where there there are multiple target domains, and the target domains are not observable during the training time. This would make our work have wider application.

% Optionally include extra information (complete proofs, additional experiments and plots) in the appendix.
% This section will often be part of the supplemental material.

\end{document}